\documentclass[10pt,twocolumn,letterpaper]{article}

\usepackage{iccv}
\usepackage{times}
\usepackage{epsfig}
\usepackage{graphicx}
\usepackage{amsmath}
\usepackage{amssymb}
\usepackage{comment}
\usepackage{color}
\usepackage{url}
\usepackage{booktabs}
\usepackage[ruled,linesnumbered]{algorithm2e}
\usepackage{dsfont}
\usepackage{wrapfig}
\usepackage{xcolor}
\usepackage{soul}
\usepackage{tikz}
\usetikzlibrary{external} 
\usetikzlibrary{matrix, calc, positioning, shapes, shadows, arrows, fit, automata}

\newcommand{\topic}[1]{\textit{\textbf{#1}}}

\usepackage[pagebackref=true,breaklinks=true,letterpaper=true,colorlinks=true,linkcolor=red, citecolor=teal,bookmarks=false]{hyperref}

\iccvfinalcopy % *** Uncomment this line for the final submission

\def\iccvPaperID{****} % *** Enter the ICCV Paper ID here

\ificcvfinal\fi

\begin{document}

%%%%%%%%% TITLE
\title{DELO: Deep Evidential LiDAR Odometry using Partial Optimal Transport}
\vspace{-0.4cm}
\author{Sk Aziz Ali$^{\star\dagger}$
%\\
%{\tt\small sk\_aziz.ali@dfki.de}
\and
Djamila Aouada$^{\dagger}$
%\\
%{\tt\small djamila.aouada@uni.lu}
\and
Gerd Reis$^{\star}$
%\\
%{\tt\small reis@dfki.de}
\and
Didier Stricker$^{\star}$
%\\
%{\tt\small didier.stricker@dfki.de}
\and
$^{\dagger}$SnT, University of Luxembourg\\
\and
$^{\star}$German Research Center for Artificial Intelligence (DFKI)
}

\maketitle
%\ificcvfinal\thispagestyle{empty}\fi

%%%%%%%%% ABSTRACT
\begin{abstract}
Accurate, robust, and real-time LiDAR-based odometry (LO) is imperative for many applications like robot navigation, globally consistent 3D scene map reconstruction, or safe motion-planning. Though LiDAR sensor is known for its precise range measurement, the non-uniform and uncertain point sampling density induce structural inconsistencies. Hence, existing supervised and unsupervised point set registration methods fail to establish one-to-one matching correspondences between LiDAR frames. We introduce a novel deep learning-based real-time ($\sim$35-40ms per frame) LO method that jointly learns accurate frame-to-frame correspondences and model's predictive uncertainty (PU) as evidence to safe-guard LO predictions. In this work, we propose (i) partial optimal transportation of LiDAR feature descriptor for robust LO estimation, (ii) joint learning of predictive uncertainty while learning odometry over driving sequences, and (iii) demonstrate how PU can serve as evidence for necessary pose-graph optimization when LO network is either under or over confident. We evaluate our method on KITTI dataset and show competitive performance, even superior generalization ability over recent state-of-the-art approaches. Source codes are available.
\end{abstract}

%%%%%%%%% BODY TEXT
\vspace{-0.3cm}
\section{Introduction} 
\begin{figure*}
\centering
\includegraphics[width=0.99\linewidth]{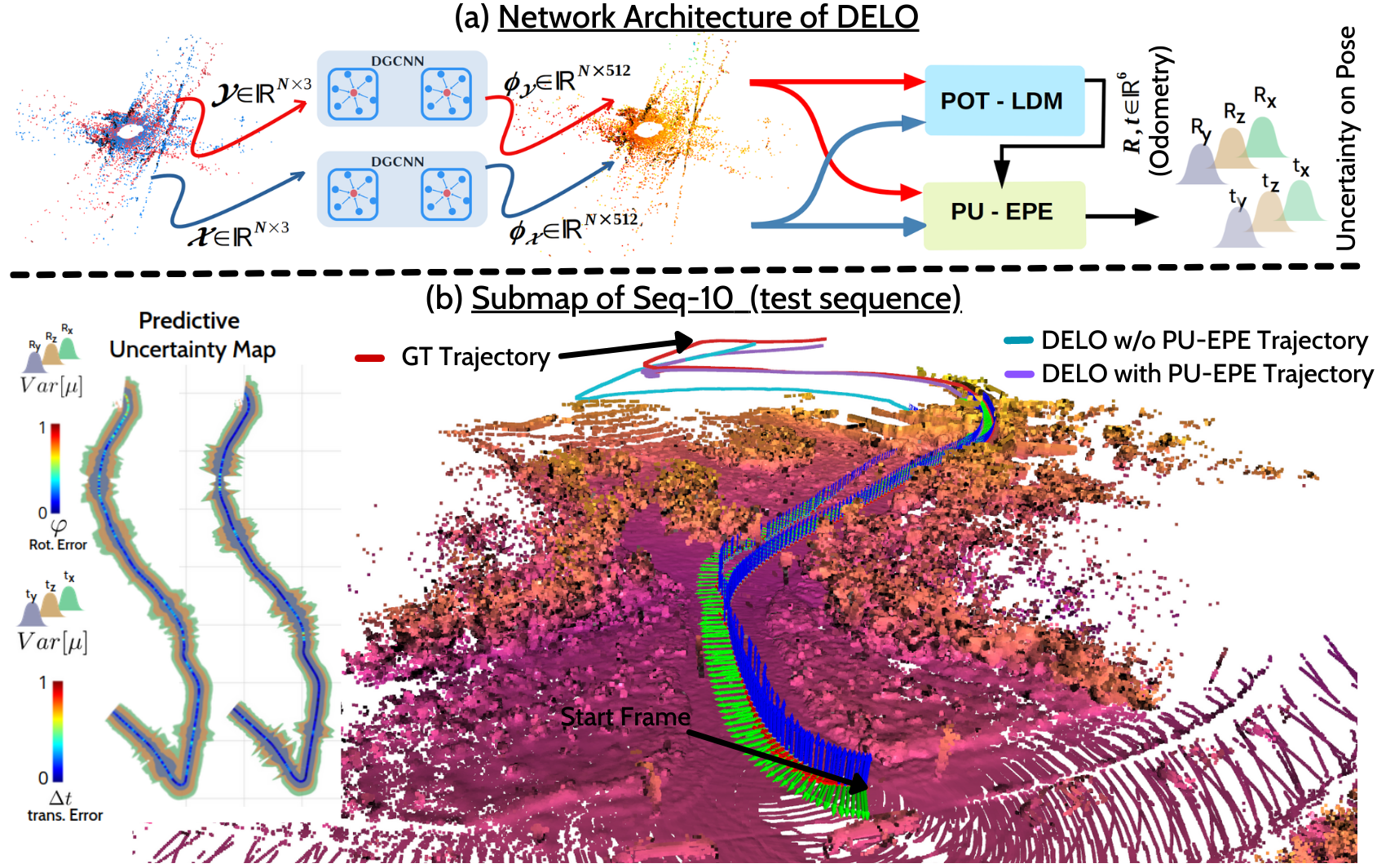}
\vspace{-0.01cm}
\caption{\topic{(a)} \textbf{Overview of DELO Network:}
Given sequential point clouds of input frames at different positions, DELO applies DGCNN \cite{WgICCV19} as backbone encoder to obtain a point-wise feature embedding. Then it simultaneously aligns the frames using the Partial Optimal Transport plan for LiDAR Descriptor Matching (POT-LDM), and estimates the Predictive Uncertainty for the Evidential Pose Estimation (PU-EPE). With the help of PU estimates pose-graphs are refined.
\topic{(b)} 
This part depicts a sub-map between frame 1 to 300 of KITTI test sequence-10 with all outputs from DELO.
}
\label{fig:Ch6_DELO_Overview}
\end{figure*}
In partial scan-to-scan alignment setting, 
LiDAR odometry (LO) is defined as the problem of estimating the 6-DoF ego-motion $\mathbf{T}_{f} \in \text{SE}(3)$, \ie,~the pose of the LiDAR sensor at frame $f$ relative to the pose at previous frame,
given two consecutive \textit{undistorted} scans.
This step serves as the backbone of most methods for robotic motion/path planning~\cite{phillips2021deep}, navigation~\cite{LiuICRA21}, simultaneous localization and mapping (SLAM)~\cite{suma2}, and many other complex scene reconstruction~\cite{weiss2011plant} tasks.

Unlike inertial or wheel odometry~\cite{ZhangAR16} using IMU sensors, LiDAR-only odometry estimation is more challenging. 
This is due to four primary reasons -- \topic{(1)} non-uniform point sampling \textit{density} of the sensor induces structural imbalance into the scan, \topic{(2)} change of speed in the moving sensors results into an out of order distribution (OOD) of relative sensor motion, \topic{(3)} scanned points that are acquired by LiDAR sensor in consecutive frames include large number of false positives and \textit{uncertain} matching correspondences, and finally \topic{(4)} previous three factors \ie~\textit{density, distribution, and uncertainty} alleviate the solution multiplicity~\cite{faugeras1990} problem significantly. 
For these reasons, the challenges in estimating relative motion of LiDAR sensor are greater than general rigid point set registration (RPSR) methods. Primarily, inconsistent drift or velocity model of ego-vehicle, dynamic objects in the scene, and cumulative error propagation due to susceptible pose predictions of all the intermediate frames are the extra difficulties for LO methods.
For these reasons, robust LiDAR point correspondence matching and self-supervised LO estimation is still an open problem. 

In this paper, we propose a Deep Evidential LiDAR Odometry (DELO) to overcome the aforementioned challenges using a unified multi-task learning approach (see Figure~\ref{fig:Ch6_DELO_Overview}). In summary, our main \textbf{contributions} are -- \topic{(i)} designing a neural network for frame-to-frame LiDAR descriptor matching (LDM) (Sec.~\ref{DELO:POT_Matching}) using partial optimal transport (POT)~\cite{RemiMLR21,Villani09,CuturiNIPS13} plan, termed as POT-LDM. This network assigns a higher weights to inliers, even if they are small in numbers between two LiDAR frames. This is a natural way to tackle of the solution multiplicity~\cite{faugeras1990} problem in LiDAR feature matching. 
Next, \topic{(ii)} a network that learns predictive uncertainty for evidential pose estimation, termed as PU-EPE (Sec.~\ref{DELO:Deep_Evidence}). We describe how the learned uncertainty over predicted poses are approximately equivariant along different transformation axes. This is an elegant way to classify under-confident, confident, and over-confident LO predictions.  Finally, \topic{(iii)} we show how the pose uncertainty can act as evidences of anomaly related to LO estimation. Herein, the dynamic pose refinement (Sec.~\ref{DELO:PoseGraphOpt}) prevents further propagation of prediction errors.   

\section{Related Work}\label{DELO:RelatedWork}
\vspace{-0.1cm}
\noindent\textbf{Deep Learning-based Point Set Registration.}
A set of purely geometric rigid point set registration (RPSR) algorithms \cite{AliACCESS21,AliCVPR21,WgICCV19,ChoyCVPR20,KeRAL21,ZhangAR16,lu2019deepvcp,ShanIROS18} can be used for LO estimation. Among them, only a few methods~\cite{AliCVPR21,ChoyCVPR20,liu2021balm,lu2019deepvcp,WgICCV19,wang2019prnet} propose deep-learning based approaches, and even fewer methods \cite{AliCVPR21,WgICCV19} can infer in real-time. Above all, inhomogeneous distribution of LiDAR points raises a common problem for all LO and registration methods to find one-to-one matching correspondences.
For instance, ICP~\cite{BlMcTPAMI92}, ICP$_{\perp}$~\cite{segal2009generalized}, CPD~\cite{MoTPAMI10}, and other classical approaches \cite{GolCVPR16,ZhECCV16} for RPSR, except FGA \cite{AliACCESS21}, all perform poorly on LiDAR scans. 
More recently, the deep neural network (DNN) for point set registration -- DCP \cite{WgICCV19}, RPSRNet \cite{AliCVPR21}, DeepVCP \cite{lu2019deepvcp} and DGR \cite{ChoyCVPR20}, appear as benchmarks for frame-to-frame registration of LiDAR scans. These methods \cite{WgICCV19,AliCVPR21,lu2019deepvcp,ChoyCVPR20} are faster and perform better than classical techniques \cite{GolCVPR16,AliACCESS21,ZhECCV16,BlMcTPAMI92}. Where RPSRNet (with $\sim$20ms inference speed) proposes a novel hierarchical representation for inhomogeneous point cloud data, DGR (with $\sim$700ms inference speed) combines a compact geometric feature map with weighted Orthogonal Procrustes (OP) \cite{gower1975generalized} for effective correspondence matching. DCP~\cite{WgICCV19}, a neural version of ICP \cite{BlMcTPAMI92}, is the first method to use transformer network \cite{VaswaniNIPS17}. It computes a \lq doubly stochastic cross-attention score matrix\rq~to find correspondences between two scans. Despite impressive formulation and learning strategy of DCP and its successor PRNet~\cite{wang2019prnet}, both suffers from well-known solution-multiplicity~\cite{faugeras1990}, \ie,~spurious association between false feature correspondences. 

\vspace{0.15cm}
\noindent\textbf{Deep Learning-based LiDAR Odometry.}
Classical ICP-based methods~\cite{zhang2014loam,shan2018lego,CT-ICP2022,ji2019lloam} and few carefully designed deep-learning-based methods \cite{PWCLO-Net,DeepLOUnsprvKr2020,Li_2019_CVPR,DeLORAICRA21,DeepLO_DMLO} span the baselines for LiDAR odometry. In our knowledge, we see many of these methods \cite{Li_2019_CVPR,DeepLO_DMLO,DeepLOUnsprvKr2020,CT-ICP2022,ji2019lloam,zhang2014loam,shan2018lego} convert 3D LiDAR scans to 2D range images and therefore undermine the vertical pose-drift by scaling it only to few pixels. LO-Net~\cite{Li_2019_CVPR}, PWCLONet~\cite{PWCLO-Net}, and RPSRNet~\cite{AliCVPR21} are among the learning-based real-time methods that directly operate on 3D point clouds. LO-Net uses point normal vectors for \lq local geometric consistency\rq~and additional mapping network module (\ie,~scan-to-map registration) for refined odometry estimation. Similar idea also exists in unsupervised learning \cite{DeepLOUnsprvKr2020,DeLORAICRA21}. Instead of relying on geometrical features, PWCLONet demonstrates how to hierarchically build a feature pyramid of point motion~\cite{LiuFlowNetCVPR19} between two scans. The main reason behind such choice is to filter small relative motions between dynamically moving objects (\ie,~scene-flow) and capturing large ego-motion for odometry. This technique effectively avoid solution-multiplicity~\cite{faugeras1990}. LO-Net (with mapping), RPSRNet, and PWCLONet (with $\sim$80ms, $\sim$20ms, and $\sim$125ms inference speed respectively) all have reported low drift compared to LOAM~\cite{zhang2014loam}(without mapping). 
%
%
%
%

% \vspace{0.15cm}
% \noindent\textbf{LO for Localization, Re-localization, and Mapping.}
% A robust LO method cannot solve other parts of SLAM problems that are equally important for globally-consistent scene and trajectory reconstruction.

\vspace{0.15cm}
\noindent\textbf{LO Uncertainty as Evidence.}
For an end-to-end trainable odometry network, it is difficult to define a direct map that captures a small perturbation in its measurement (\ie,~$\mathbf{T}_{f}$) at the current frame $f$ and readjust the changes for the future measurements (\ie,~$\mathbf{T}_{f+1},\hdots$). This problem is occurs due to unexpected out-of-order distribution (OOD). Therefore, \textit{uncertainty quantification (UQ) over the predicted transformations helps in setting boundary conditions for any downstream pose-based decision making tasks} \cite{ABDAR2021243,KendallNIPS17} -- \eg,~general classification \cite{sensoy2018evidential}, motion forecasting \cite{phillips2021deep}, and navigation \cite{LiuICRA21}. A recent deep multi-task learning approach LP2 \cite{phillips2021deep}, for joint localization, perception and prediction tasks underpins the importance of UQ in their setup. Deep Evidential Regression (DER) \cite{AminiNIPS20} is now a preferred choice for many learning-based navigation models \cite{LiuICRA21,sensoy2018evidential} than conventional and computationally costly UQ techniques \cite{BlasnAWS08,ABDAR2021243}. To this end, joint learning of LO and PU, and thereafter using such relational model for automatic odometry refinement, remain unexplored.

% To refine LO estimates, a network for predictive uncertainty (PU) estimation is associated as evidence. The proposed framework jointly learns the PU and LO, and reuses their shared weights as an actuator pose-graph optimization. Extensive evaluation shows that our method outperforms state-of-the-art LO methods.
 
%
%
%
\vspace{-0.2cm}
\section{DELO Method Overview}
\label{DELO:Architecture}
\vspace{-0.15cm}
The proposed DELO operates on a sequence of 3D LiDAR scans \hbox{$\mathbf{\mathcal{S}}=\lbrace\mathbf{\mathcal{X}}_f\rbrace_{f=1}^{S}$}, where 
any input scan $\mathbf{\mathcal{X}}_f$ at frame $f$ is randomly sub-sampled to a fixed $N$ number of points. 
Thereafter, DELO takes pair-wise source $\mathbf{\mathcal{Y}} \in \mathbb{R}^{N\times 3}$ and target $\mathbf{\mathcal{X}}\in \mathbb{R}^{N\times 3}$ point clouds as input, and embeds them independently with DGCNN \cite{WangTOG19} encoder $\phi: \mathbb{R}^{N\times3} \rightarrow \mathbb{R}^{N\times D}$. We have used $N=1024$, $D=512$. 
% For each pair of consecutive scans, the features are made context specific by a transformer network \cite{VaswaniNIPS17} resulting in a 1024-dimensional feature per point $\boldsymbol{\Phi} \in \mathbb{R}^{N\times1024}$.

\vspace{-0.1cm}
\subsection{POT-LDM: Sharp Correspondence Matching} %-- Partial Optimal Transport Plan}
\label{DELO:POT_Matching}
\begin{figure}[ht]
\centering
\includegraphics[width=0.99\linewidth]{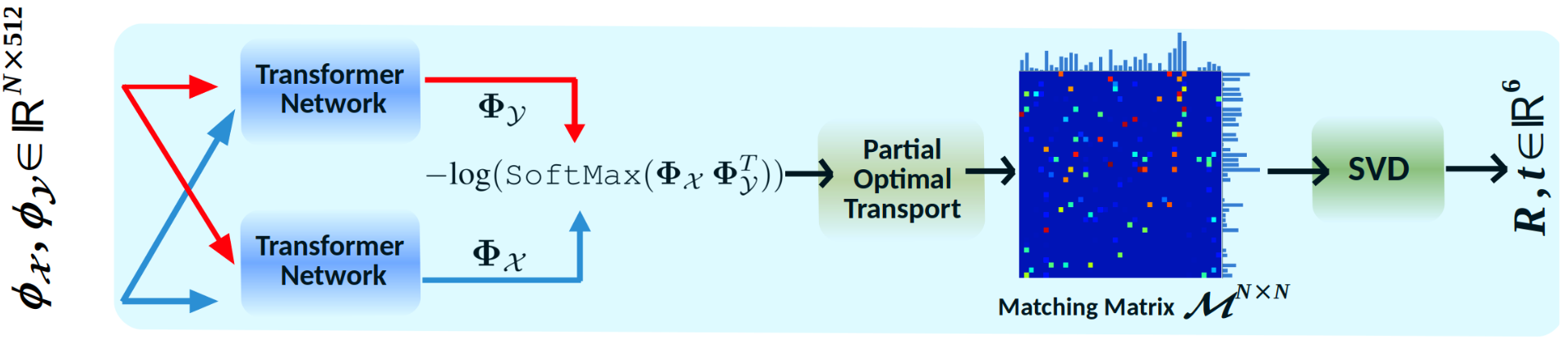}
\vspace{-0.1cm}
\caption{POT-LDM network architecture.}
\label{fig:Ch6_DELO_POT}
\end{figure}
The POT-LDM network, as shown in Figure~\ref{fig:Ch6_DELO_POT}, takes point-wise feature embedding $\mathbf{\phi}_{\mathbf{\mathcal{Y}}} \in \mathbb{R}^{N\times D}$ and
$\mathbf{\phi}_{\mathbf{\mathcal{X}}} \in \mathbb{R}^{N\times D}$ of source $\mathbf{\mathcal{Y}}$ and target $\mathbf{\mathcal{X}}$. 
Next, a transformer network \cite{VaswaniNIPS17} turns the input
features into task-specific features \cite{WgICCV19} using contextual map
$\varphi: \mathbb{R}^{N\times D} \times \mathbb{R}^{N\times D} \rightarrow \mathbb{R}^{N\times D}$. This is an asymmetric learnable map for measuring changes between two input embedding tensors 
$\phi_{\mathcal{Y}}$ and $\phi_{\mathcal{X}}$. Finally, the output of the transformer network, 
\vspace{-0.2cm}
\begin{equation}\label{eqn:Attention_DELO}
\boldsymbol{\Phi}_{\mathcal{Y}} = \boldsymbol{\phi}_{\mathcal{Y}} + \varphi(\boldsymbol{\phi}_{\mathcal{Y}}, \boldsymbol{\phi}_{\mathcal{X}})
\,\text{and}\, 
\boldsymbol{\Phi}_{\mathcal{X}} = \phi_{\mathcal{X}} + \varphi(\boldsymbol{\phi}_{\mathcal{X}}, \boldsymbol{\phi}_{\mathcal{Y}}),
%\vspace{-0.2cm}
\end{equation}
are used for sharp correspondence matching. For this, DCP~\cite{WgICCV19} runs differentiable
soft-assignment such that for each point \hbox{$y_i\in\mathcal{Y}$}, a probability vector over 
$\mathcal{X}$ is assigned as matching measure $m(y_i, \mathcal{X}) = \texttt{SoftMax}(\mathbf{\Phi}_{\mathcal{X}} \mathbf{\Phi}_{y_{i}}^{T})$.

\vspace{0.1cm}
\noindent\textbf{Partial Transportation of Mass as Attention Weights.}
\begin{figure*}[ht]
\centering
\includegraphics[width=0.99\linewidth]{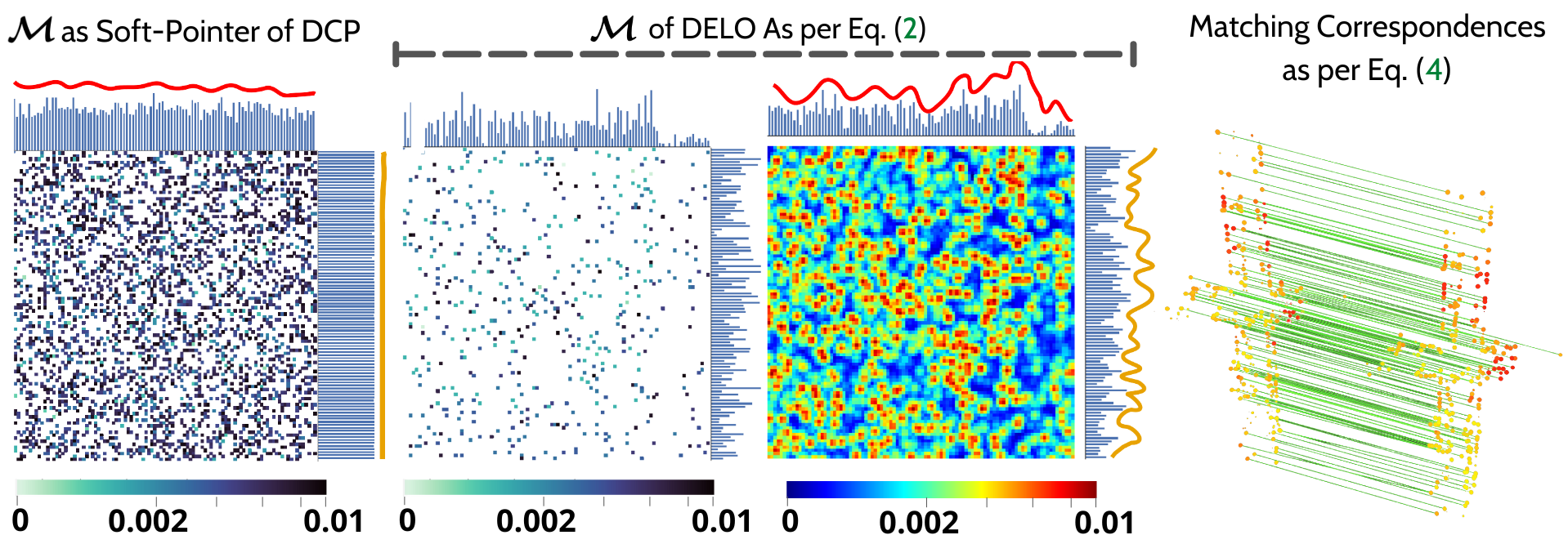}
\caption{\textbf{Matching Matrix Comparison:} DCP \cite{WgICCV19} Soft-Pointer based
matching matrix $\boldsymbol{\mathcal{M}}$ (on the left). The histogram plot over the rows and columns 
denote sum of probability vectors along the dimensions. Two plots from the right (with different
colormaps for visual clarity) denote $\boldsymbol{\mathcal{M}}$ when optimized by POT-LDM.}
\label{fig:Ch6_MatchingMatrix_Comparison}
\end{figure*}
\noindent Due to sensor movement and sparse point clouds in LiDAR-data, we observe only a partial number of points that can be matched between consecutive scans. While point set registration only requires a minimum of three true point correspondences for solving OP \cite{gower1975generalized} problem, it is difficult to know or match true feature correspondences in advance. 
Therefore, one can assume this as a partial-to-partial sparse rigid point set registration task. We employ partial optimal mass transportation technique \cite{Villani09}
for \textit{sharp} point matching instead of its seminal version \cite{CuturiNIPS13}. In this technique, low transportation cost means input features match closely. We use entropy regularized partial optimal transport~\cite{BenSIAM15} 
\begin{equation}
  \begin{aligned}
      \boldsymbol{\mathcal{M}} = \arg \min_{\boldsymbol{\mathcal{M}}}
      \left\langle \boldsymbol{\mathcal{M}}, \boldsymbol{C}\right\rangle_F 
      + 
      \lambda \Omega(\boldsymbol{\mathcal{M}}),\\
      \text{s.t.}\,\, \boldsymbol{\mathcal{M}} \boldsymbol{1} \leq a,\\
      \boldsymbol{\mathcal{M}}^T \boldsymbol{1} \leq b,\,\text{and}\\
      \boldsymbol{1}^T \boldsymbol{\mathcal{M}}^T \boldsymbol{1} = m \leq \min\{a^T\boldsymbol{1}, b^T\boldsymbol{1}\},
  \end{aligned}
\end{equation}
where $\left\langle\boldsymbol{\mathcal{M}}\,,\boldsymbol{\mathcal{C}}\,\right\rangle_{F}=\text{tr}\left(\boldsymbol{\mathcal{M}}^{\text{T}}\boldsymbol{\mathcal{C}}\right)$ denotes Forbenius norm over matrix dot product, $\boldsymbol{1}=(1, \hdots,1)^{T}$ is a vector of all $N$ elements as 1, $\boldsymbol{\mathcal{M}} \in (\mathbb{R}_+)^{N\times N}$ is the transport matrix, 
$\boldsymbol{C} \in (\mathbb{R}_+)^{N\times N}$ is the cost matrix, $\lambda$ 
is a regularization parameter, $a \in \mathbb{R}_{+}^{N\times1}$ and $b \in \mathbb{R}_{+}^{N\times1}$ are probability distributions, $m$ 
is mass to transport and 
$\Omega (\boldsymbol{\mathcal{M}}) = \sum_{i,j} \boldsymbol{\mathcal{M}}_{i,j} 
\text{log}(\boldsymbol{\mathcal{M}}_{i,j})$ 
is the entropic regularization term. 
The amount of transported mass $m$ 
between both inputs acts as a control parameter to adjust the \lq sharpness\rq ~of the correspondence matching. 
The regularization parameter $\lambda$ is learned during network training. 
Each point initially has an equal probability of $a,b= \boldsymbol{1} / N$. 
We set the cost matrix $\boldsymbol{C}$ as the negative log-likelihood of the matching probabilities \cite{WgICCV19} for every point $y_i\in \mathcal{Y}$ 
with all points in $\mathcal{X}$ such that its matching cost
%\vspace{-0.15cm}
\begin{equation}
  \boldsymbol{C}_{y_i} = - \text{log}(\texttt{SoftMax}(\mathbf{\Phi}_{\mathcal{X}}\,\mathbf{\Phi}_{y_i}^T)).
  \vspace{-0.1cm}
\end{equation} 
The negative log-likelihood penalizes the outlier points. 
\SetKwComment{Comment}{/* }{ */}
\RestyleAlgo{ruled}
\begin{algorithm}
    \caption{Partial Optimal Transport}\label{alg:Ch6_POT}
    \KwData{Cost matrix $\boldsymbol{C}$, mass m, Iter. $\xi$, regularizer $\lambda$}
    \KwResult{Partial Optimal Transport Mass $\boldsymbol{\mathcal{M}}$}
    \Begin{
        $a, b \gets \mathbf{1}/N \, \text{and}\, \mathbf{1}/N$\;
        $\boldsymbol{K} \gets e^{-\boldsymbol{C}/\lambda}$\;
        \For{$i \gets$ \KwTo $\xi$}{
            $\boldsymbol{\tilde{K}} \gets \text{diag}(\min(\frac{a}{\boldsymbol{\tilde{K}}\mathbf{1}}, \mathbf{1}))\boldsymbol{K}$\;
            $\boldsymbol{\hat{K}} \gets \boldsymbol{\tilde{K}} \text{diag}(\min(\frac{b}{\boldsymbol{\tilde{K}}^T\mathbf{1}}, \mathbf{1}))$\;
            $\boldsymbol{K} \gets \boldsymbol{\hat{K}}\frac{m}{\mathbf{1}^T \boldsymbol{\hat{K}}\,\mathbf{1}}$\;
        }
        $\boldsymbol{\mathcal{M}} \gets \boldsymbol{K}$\;
    }
\end{algorithm}
Algorithm \ref{alg:Ch6_POT} describes entropy regularized POT solution steps following
\cite{BenSIAM15} and \cite{RemiMLR21}. It is possible to set different mass initialization 
and iteration limits to the input of Algorithm \ref{alg:Ch6_POT}. An optimal transportation cost is achieved faster and efficiently if one initially sets low masses $m$ and a fewer number of iterations for Algorithm \ref{alg:Ch6_POT}. The scores from the matching matrix 
$\boldsymbol{\mathcal{M}}$
% (alternatively termed as Optimal Transport Mass matrix)
are used to compute the rigid transformation $\mathbf{T} = \left[\mathbf{R}, \mathbf{t}\right]$ 
using weighted-Procrustes \cite{YewCVPR20,ChoyCVPR20,CattToR22} with differentiable SVD. 

\vspace{0.15cm}
The proposal of local sharp matching through differentiable 
POT module reduces matching cost between the attention maps $\mathbf{\Phi}_{\mathcal{Y}}$ 
\text{and} $\mathbf{\Phi}_{\mathcal{X}}$. This is more effective way than the soft-pointer driven feature matching in DCP \cite{WgICCV19}. The resulting transportation or matching matrix $\boldsymbol{\mathcal{M}}$ from Algorithm \ref{alg:Ch6_POT} produces  
smaller number of matched features, but \lq sharper\rq~in probabilities matches compared to the soft-pointer
approach of \cite{WgICCV19}. The first plot in Figure~\ref{fig:Ch6_MatchingMatrix_Comparison}
with resolution\footnote{Note that the original cost matrix $\boldsymbol{C}$ 
has dimension $1024 \times 1024$. We applied stride convolution with filter size $10\time 10$ to 
lower its resolution for visual purpose. Hence the cost entries are only marginally scaled.} $102 \times 102$
explains that for any given source point $y_i \in \mathcal{Y}$ (along the column), its total matching 
probabilities with all other target points (along the row) are approximately constant. 
On the other hand, the other two plots in the same figure show how the same cost distribution, after optimizing $\boldsymbol{C}$  (setting $m=0.1$ and $\xi=5$ in Alg. \ref{alg:Ch6_POT}),  are optimally transported by $\boldsymbol{\mathcal{M}}$.

\vspace{0.1cm}
\noindent\textbf{6DoF Pose Regression Loss.} After estimating the optimal $\boldsymbol{\mathcal{M}}$, every point from the source point cloud 
$y_i \in \mathbf{\mathcal{Y}}$ is mapped to the location 
\vspace{-0.25cm}
\begin{equation}\label{eqn:Correspondence_Projeciton_M}
 \tilde{y}_i \gets \frac{1}{\sum_{j=1}^{N}\boldsymbol{\mathcal{M}}_{ij}}  \sum_{j=1}^{N} \boldsymbol{\mathcal{M}}_{ij} y_i
\end{equation}
that corresponds to its target position $\tilde{y}_i$.
% Next, differential weighted SVD module for WOPP (see Sec. \ref{Chapter3:RigidMotionField_in_RGBD_Scene_and_LiDAR_Scan_Registration:EstimatingSensorPoses:Procrustes}),
% as used by DGR method \cite{ChoyCVPR20}, estimates the rigid transformation $\mathbf{T}$.  
The loss function for sharp local feature matching-based pose estimation is a combination of 
pose loss $\mathcal{L}_{pose}$ and an auxiliary loss $\mathcal{L}_{aux}$ (as referred to by \cite{CattToR22}). The pose loss is the $\ell_{1}$-norm between the source points (\ie,~LiDAR points of current frame) transformed by the ground truth transformation
$\mathbf{T}_{\text{gt}}$ and the predicted transformation $\mathbf{T}$:
\vspace{-0.15cm}
\begin{equation}
  \label{eq:pose}
  \begin{aligned}
    \mathcal{L}_{pose} = \frac{1}{N} \sum_{i=1}^{N} \left\lvert\mathbf{T}_{\text{gt}}y_i - \mathbf{T}y_i\right\rvert \, \text{and} \,
    \mathcal{L}_{aux} = \frac{1}{N} \sum_{i=1}^{N} \left\lvert \tilde{y}_i - \mathbf{T}y_i\right\rvert. 
  \end{aligned}
\end{equation} 
\vspace{-0.15cm}
The total odometry loss is defined as
\begin{equation}\label{eqn:total_Reg_Loss}
 \mathcal{L}_{odom} = \mathcal{L}_{pose} + \lambda_{aux}\mathcal{L}_{aux} \,\,\text{where}\,\, \lambda_{aux}=0.05.
\end{equation}
% where the default value of 
% $\lambda_{aux}$ is set to $0.05$.

\subsection{PU-EPE: Evidential Pose Estimation}\label{DELO:Deep_Evidence}
\begin{figure}[ht]
\centering
\includegraphics[width=0.99\linewidth]{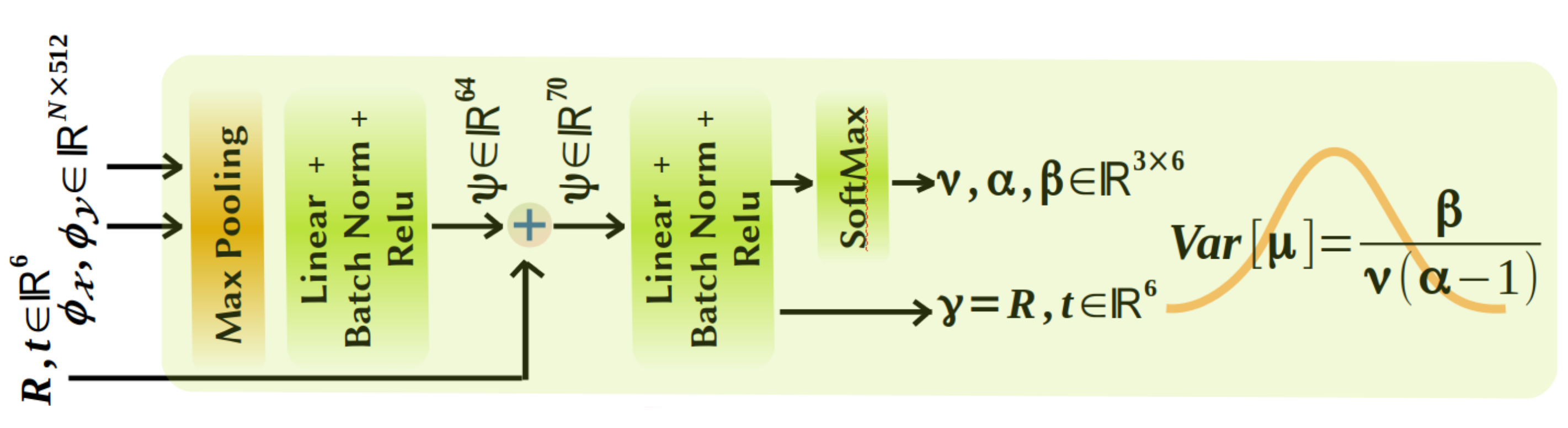}
\vspace{-0.1cm}
\caption{The PU-EPE network takes the embedding vectors $\mathbf{\phi}_{\mathcal{Y}}$ and $\mathbf{\phi}_{\mathcal{X}}$  of source and target frames, and predicts both aleatoric 
and epistemic uncertainty parameters~\cite{AminiNIPS20}.}
\label{fig:Ch6_DELO_PUEPE}
\end{figure}
%
%
% The second block of the DELO architecture is a novel evidence module, named as predictive 
% uncertainty-based evidential pose estimation (PU-EPE).
%
%
\vspace{-0.15cm}
Once optimal correspondence association is established, measurement noise is reduced. Though, out-of-order (OOD) data distribution can still induces uncertainties in the next steps. To overcome this problem, aggressive data augmentation can be one option for a network to learn such OOD. While the POT-LDM can inherently learn the data uncertainty (\ie, aleatoric uncertainty), it cannot automatically learn the model's predictive uncertainty (\ie, epistemic uncertainty) \cite{KendallNIPS17}. 
The PU-EPE network, as shown in Figure~\ref{fig:Ch6_DELO_PUEPE}, is trained jointly with the POT-LDM network to estimate the neural model's confidence in the LO predictions at different frames, \ie, $\mathbf{T}_f, \mathbf{T}_{f+o}, \mathbf{T}_{f+2o},$ .., when frame-gap is $o$.
% \vspace{0.1cm}
In theory, the PU-EPE model learns to maximize the negative log-likelihood of the observed transformation values $\mathbf{T}_{f} \in \mathbb{R}^6$ that are assumed to be drawn from an independent and identically distributed (i.i.d) realization of a Gaussian distribution 
with unknown mean $\mu$ and variance $\sigma^2$. To learn the epistemic uncertainty parameters, the actual distribution mean $\mu$ and variance $\sigma^2$ are estimated by inferring the hyper-parameters $\gamma, \nu, \alpha$ and $\beta$ of a 
Normal-Inverse-Gamma (NIG) distribution $p(\mu, \sigma^2 | \gamma, \nu, \alpha, \beta)$ 
\begin{equation}
   = \frac{\beta^{\alpha} \sqrt{\nu } }{ \Gamma (\alpha) \sqrt{2\pi \sigma^2}}  (\frac{1}{\sigma^2})^{\alpha+1}  exp \{  - \frac{2 \beta + \nu  (\gamma - \mu )^2}{2 \sigma^2} \}
\end{equation}
as a posterior distribution\cite{AminiNIPS20} of the NIG.

\vspace{0.1cm}
By drawing i.i.d samples from the above NIG distribution, the PU-EPE model can directly infer 
both the epistemic uncertainty as $\text{Var}[\mu] = \frac{\beta}{\nu (\alpha - 1)}$ and the
aleatoric uncertainty as $\mathbb{E}[\sigma^2] = \frac{\beta}{\alpha - 1}$, and $\mathbb{E}[\mu] = \gamma$, 
without undertaking costly sampling~\cite{BlasnAWS08} techniques. To infer the hyper-parameters of 
the NIG distribution, the transformation $\mathbf{T} \in \mathbb{R}^{6}$ predicted by the POT-LDM network is concatenated with the output of a linear feed-forward multi-layer perceptron (MLP) applied on  $\phi_{\mathcal{Y}}$ and $\phi_{\mathcal{X}}$.
% This behaves as positional encoding  with context-based information. 
Finally, the hyper-parameters 
are inferred using a second MLP as shown by green block \lq Linear + Batch Norm. + ReLu\rq. Following Amini \etal~\cite{AminiNIPS20}, we define two losses namely, the negative log likelihood loss
\vspace{-0.3cm}
\begin{equation}
  \label{eq:evidence_nll}
  \begin{aligned}
      \mathcal{L}_{NLL}^k = \frac{1}{2} log (\frac{\pi}{\nu_k}) - \alpha_k \text{log}(\Omega_k)\\
      + (\alpha_k + \frac{1}{2}) \text{log}((\mathbf{T}_k - \gamma_k)^2 \nu_k + \Omega_k)
      + log(\frac{\Gamma(\alpha_k)}{\Gamma(\alpha_k+ 1/2)})
  \end{aligned}
\end{equation}
that minimizes the evidence on transformation errors, and the regularization loss
\vspace{-0.15cm}
\begin{equation}
  \label{eq:evidence_reg}
  \mathcal{L}_{R}^k = \left\lvert \mathbf{T}_k - \gamma_k \right\rvert * (2 \alpha_k + \nu_k)
  \vspace{-0.1cm}
\end{equation}
that maximizes the model fitting, where $\Omega_k = 2\beta_k(1 + \nu_k)$, and the subscript $k$ 
for all hyper-parameters $\gamma_k, \nu_k, \alpha_k, \beta_k$ indicates the
$k^{\text{th}}$ element out of 6 (3 for Euler angles and 3 for translation) DoF. The total evidence loss
\vspace{-0.15cm}
\begin{equation}\label{eqn:total_evidential_loss}
  \mathcal{L}_{evidence} = \frac{1}{6} \sum_{k = 1}^{6} (\mathcal{L}_{NLL}^k + \lambda_R\mathcal{L}_{R}^k)
  \vspace{-0.1cm}
\end{equation}
is mean loss over all 6 transformation parameters. The regularization parameter 
$\lambda_R$ is set to 0.2 for both the rotation and the translation parameters.
%
%
% In the end, the both the odometry loss $\mathcal{L}_{odom}$ and evidence loss $\mathcal{L}_{evidence}$ are jointly learned. 
%
%
The uncertainty in the model's LO predictions is used in the later stages of our method.

\vspace{-0.1cm}
\subsection{Pose Refinement using Uncertainty as Evidence}
\label{DELO:PoseGraphOpt}
\vspace{-0.1cm}
%
%
% \begin{figure}[ht]
% \centering
% \includegraphics[width=0.99\linewidth]{figures/PoseGraphOpt.png}
% \vspace{-0.2cm}
% \caption{Frame-factor adjustments for pose-graph optimization and loop detection \hbox{w.r.t} the frame-factor of LiDAR odometry}
% \label{fig:Ch5_PoseGraphOPtimization}
% \end{figure}
%
%
\begin{figure}[ht]
\vspace{-0.3cm}
\centering
    \includegraphics[width=0.47\textwidth]{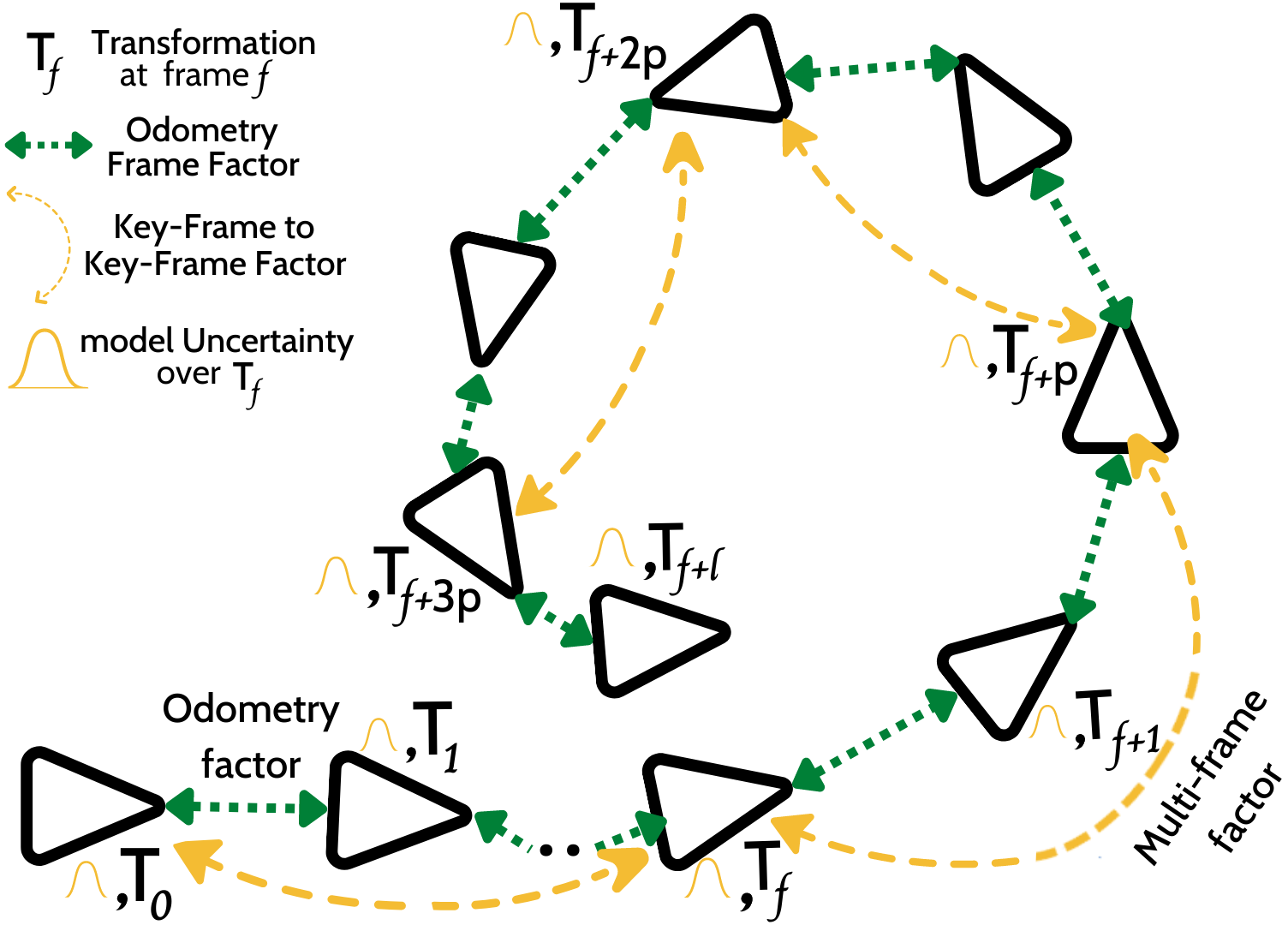}
    \vspace{-0.3cm}
    % \caption{Frame-factor adjustments for pose-graph optimization \hbox{w.r.t} the frame-factor of odometry}
\label{fig:Ch5_PoseGraphOPtimization}
\end{figure}
In the final functional component of DELO, the odometry predictions  from POT-LDM $\left[..,\mathbf{T}_f, \mathbf{T}_{f+o}, \mathbf{T}_{f+2o}, \hdots\right]$, model's or epistemic uncertainty measures $[..,\text{Var}\left[\mu \right]_f,..]$ 
%
%
% \begin{wrapfigure}{r}{0.38\textwidth}
% \vspace{-0.4cm}
% \centering
%     \includegraphics[width=0.38\textwidth]{figures/PoseGraphOpt}
%     \vspace{-1.0cm}
%     % \caption{Frame-factor adjustments for pose-graph optimization \hbox{w.r.t} the frame-factor of odometry}
% \label{fig:Ch5_PoseGraphOPtimization}
% \end{wrapfigure}
%
%
from PU-EPE for all frames $f$ in a given sub-sequence $\mathbf{\mathcal{S}}$ are all streamed in for pose-graph optimization. Two parallel threads can run two tasks separately on different frame factors. The figure above depicts how one key-frame to another key-frame factor $p$ for pose-graph optimization, and odometry frame-factor $o$ can be set. Similar to some conventional methods~\cite{ShanIROS18,ramezani2020online}, we also employ GTSAM \cite{GtSam12} for pose-graph optimization by matching the LO inference with LiDAR sensor rate. 
The orange curve in the figure shows factor graphs \cite{KschTIT01} built over local frames and previous key-frames with \textit{multi-frame factors}.
The intermediate frames can be more in numbers and hierarchical as well \cite{DrochlICRA18}.
On the other hand green edges indicate frame-to-frame pose graph only up to odometry frame-factor. 
Once our DELO network is trained, we build an empty pose-graph during trajectory inference stage by adding the predicted poses $\left[\mathbf{T}_{f}, \mathbf{T}_{f+p}, \mathbf{T}_{f+2p},...\right]$. The odometry factor $\textit{o}$ and 
multi-frame pose-graph factor $p$ are set to 2 and 4 respectively. If confidence score (\ie, $1-\text{Var}\left[\mu\right]$) of DELO network for LO prediction between
two key-frames is bounded by predefined thresholds $\theta_{min}$ and $\theta_{max}$, \ie, 
\begin{equation}\label{eqn:Confidence_Bound}
    \theta_{min} \leq 1 - \text{Var}\left[\mu\right]_{f\mapsto f+p} \leq \theta_{max},
\end{equation}
then the factor graph node is rejected. This means, the network is confident and there is no need to refine the previous poses present in the current pose-graph.

\vspace{-0.2cm}
\section{Experiments and Evaluations}
\label{DELO:Experiments_Evaluation}
In this section, we present a complete and comprehensive experimental evaluation of our proposed DELO method on KITTI LiDAR odometry dataset~\cite{Geiger2013IJRR}. 

\setlength\tabcolsep{1.25pt}
\setul{0.3ex}{0.4ex}
\setulcolor{olive}
\begin{table*}[!ht]
    \centering
    \footnotesize
    \setlength{\tabcolsep}{0.6mm}
	%\hspace*{-0.05mm}
    \begin{tabular}{l|cc|cc|cc|cc|cc|cc|cc|cc|cc|cc|cc|c|}
    \toprule
    & \multicolumn{14}{c|}{Training} 
    & \multicolumn{9}{c|}{Testing or Inference}\\
    \cline{1-24}
    Method
    & 
    \multicolumn{2}{c}{00} & 
    \multicolumn{2}{c}{01} & 
    \multicolumn{2}{c}{02} & 
    \multicolumn{2}{c}{03} & 
    \multicolumn{2}{c}{04} &
    \multicolumn{2}{c}{05} &
    \multicolumn{2}{c|}{06} &
    \multicolumn{2}{c}{07} &
    \multicolumn{2}{c}{08} &
    \multicolumn{2}{c}{09} &
    \multicolumn{2}{c}{10} &
    \multicolumn{1}{c|}{(07-10)}\\
    & $r_\text{rel}$ 
    & $t_\text{rel}$ 
    & $r_\text{rel}$ 
    & $t_\text{rel}$
    & $r_\text{rel}$ 
    & $t_\text{rel}$
    & $r_\text{rel}$ 
    & $t_\text{rel}$
    & $r_\text{rel}$ 
    & $t_\text{rel}$
    & $r_\text{rel}$ 
    & $t_\text{rel}$
    & $r_\text{rel}$ 
    & $t_\text{rel}$
    & $r_\text{rel}$ 
    & $t_\text{rel}$
    & $r_\text{rel}$ 
    & $t_\text{rel}$
    & $r_\text{rel}$ 
    & $t_\text{rel}$
    & $r_\text{rel}$ 
    & $t_\text{rel}$
    & runtime\\
    \midrule
    % FGR~\cite{ZhECCV16} 
    % & xx & xx & xx & xx & xx & xx 
    % & xx & xx & xx & xx & xx & xx 
    % & xx & xx & xx & xx & xx & xx 
    % & xx & xx & xx & xx
    % % <-- 1.5 times
    % \\
    ICP$^{\dagger}$~\cite{BlMcTPAMI92} 
    & 2.99 & 6.88 & 2.58 & 11.2 & 3.39 & 8.21 
    & 5.05 & 11.1 & 4.02 & 6.64 & 1.92 & 3.97 
    & 1.59 & 1.95 & 3.35 & 5.17 & 4.93 & 10.04 
    & 2.89 & 6.93 & 4.74 & 8.91 & -NA-
    \\ 
    ICP$^{\dagger}_{\perp}$~\cite{segal2009generalized} 
    & 1.73 & 3.80 & 2.58 & 13.53 & 2.74 & 9.00 
    & 1.63 & 2.72 & 2.58 & 2.96 & 1.08 & 2.29 
    & 1.00 & 1.77 & 1.42 & 1.55 & 2.14 & 4.42 
    & 1.71 & 3.95 & 2.60 & 6.13 & -NA-
    \\
    LOAM$^{\dagger}$~\cite{ji2019lloam} 
    & 6.25 & 15.99 & 0.93 & 3.43 & 3.68 & 9.40  
    & 9.91 & 18.19 & 4.57 & 9.59 & 4.10 & 9.16  
    & 4.63 & 8.91 & 6.76 & 10.87 & 5.77 & 12.72  
    & 4.30 & 8.10 & 8.79 & 12.67 & -NA-
    \\
    LONet$^{\dagger}$\cite{Li_2019_CVPR} 
    & 0.72 & 1.47 & 0.47 & 1.36 & 0.71 & 1.52  
    & 0.66 & 1.03 & 0.65 & 0.51 & 0.69 & 1.04  
    & 0.50 & 0.71 & 0.89 & 1.70 & 0.77 & 2.12  
    & 0.58 & 1.37 & 0.93 & 1.80 & 80.1ms
    % <-- 3.5 times, 4 times
    \\
    PWCLO$^{8\dagger}$~\cite{PWCLO-Net} 
    & \ul{\textbf{0.42}} & \ul{\textbf{0.78}} & \ul{\textbf{0.23}} & \ul{\textbf{0.67}} & \ul{\textbf{0.41}} & \ul{\textbf{0.86}}  
    & \ul{\textbf{0.44}} & \ul{\textbf{0.76}} & \ul{\textbf{0.40}} & \ul{\textbf{0.37}} & \ul{\textbf{0.27}} & \textbf{0.45}  
    & \ul{\textbf{0.22}} & \ul{\textbf{0.27}} & 0.44 & 0.60 & \ul{\textbf{0.55}} & 1.26  
    & \ul{\textbf{0.35}} & \ul{\textbf{0.79}} & \ul{\textbf{0.62}} & 1.69 & -NA-
    \\
    \midrule
    ICP~\cite{BlMcTPAMI92} 
    & 2.23 & 13.4  & 8.44 & 14.9  & 4.81 & 30.4  & 2.16 & 10.5  & 1.53 & 90.7 & 1.37 & 18.7 & 1.41 & 32.3  & 0.79 & 7.85  & 1.25 & 12.6  & 5.11 & 32.3  & 2.81 & 21.1 & 3.8s
    \\
    ICP$_{\perp}$~\cite{segal2009generalized} 
    & 3.59  & 8.23 & 8.81  & 18.5  
    & 4.09 & 12.8 & 5.5 & 12.0 & 5.21 & 16.2 & 1.79 & 3.6 & 2.05  
    & 3.97 & 5.43 & 7.9 & 6.2 & 12.4  & 3.7  & 10.4 & 4.2 & 9.7 & 10.2s
    \\
    DCP~\cite{WgICCV19} 
    & 27.8 & 58.9 & 31.3 & 96.6
    & 30.6 & 66.6 & 45.5 & 74.9  
    & 74.2 & 97.8 & 23.6 & 49.7  
    & 36.7 & 83.9 & 24.9 & 44.9  
    & 29.0 & 58.7 & 31.2 & 62.0  
    & 37.1 & 81.1 & 32ms
    \\
    DGR~\cite{ChoyCVPR20} 
    & 5.95 & 35.9  & 12.8 & 53.6  
    & 7.23 & 46.3  & 7.80 & 61.0  
    & 13.3 & 56.7  & 4.69 & 33.2  
    & 21.8 & 43.5  & 5.83 & 36.4  
    & 7.81 & 37.9  & 7.74 & 43.4  
    & 6.92 & 51.7  & 682 ms
    % <-- 4.5 times, 2 times 
    \\
    RPSRNet~\cite{AliCVPR21} 
     & 1.30 & 2.39  & 1.15 & 2.83  & 0.97 & 2.61  & 1.69 & 5.53  & 2.64 & 4.68  & 1.29 & 3.38  & 2.66 & 7.81  & 3.59 & 4.99  & 0.75 & 2.07  & 0.97 & 2.30  & 2.85 & 5.88 & \ul{\textbf{20.3}}ms
    % <-- 1.5 times, 2 times
    \\
    LOAM~\cite{zhang2014loam} 
    & 34.9 & 86.5  & 12.7 & 98.7  
    & 31.1 & 95.78 & 22.8 & 92.2  
    & \ul{{\color{gray}\textbf{1.37}}} & 97.2  & 35.9 & 83.6  
    & 33.1 & 83.7  & 61.6 & 88.1  
    & 33.2 & 87.6  & 31.8 & 93.1  
    & 28.6 & 98.1  & 121ms
    \\
    PWCLO$^{1}$~\cite{PWCLO-Net} 
    & 19.5 & 30.1  & 3.41 & 7.90  
    & 12.8 & 25.9  & 43.9 & 37.2  
    & 19.9 & 22.9  & 24.1 & 34.7  
    & 13.1 & 12.1  & 12.50 & 18.0  
    & 19.1 & 30.8  & 14.7 & 21.8  
    & 19.9 & 33.9 & 77.3ms
    % <-- 3.5 times, 4 times 
    \\
    PWCLO$^{2}$~\cite{PWCLO-Net} 
    & 2.28 & 3.41  & 1.01 & 3.02  & 1.66 & 3.83  & 2.32 & \ul{{\color{gray}\textbf{1.81}}}  & 1.45 & \ul{{\color{gray}\textbf{2.04}}}  & 1.46 & 2.02  & 0.98 & 1.32  & 1.96 & 2.26  & 1.47 & 3.21  & 1.36 & 2.31  & 2.22 & 5.80 & 82.2
    % <-- 3.5 times, 4 times 
    \\
    PWCLO$^{8}$~\cite{PWCLO-Net} 
    & \ul{\textbf{0.43}} & \ul{\textbf{0.89}}  & \ul{\textbf{0.42}} & \ul{\textbf{1.11}}  & \ul{{\color{gray}\textbf{0.76}}} & \ul{{\color{gray}\textbf{1.87}}}  & \ul{\textbf{0.92}} & 1.42  & \ul{\textbf{0.94}} & \ul{\textbf{1.15}}  & \ul{{\color{gray}\textbf{0.71}}} & \ul{{\color{gray}\textbf{1.34}}}  & \ul{{\color{gray}\textbf{0.38}}} & \ul{\textbf{0.60}}  & \ul{{\color{gray}\textbf{1.00}}} & \ul{{\color{gray}\textbf{1.16}}}  & \ul{{\color{gray}\textbf{0.72}}} & \ul{{\color{gray}\textbf{1.68}}}  & \ul{\textbf{0.46}} & \ul{\textbf{0.88}}  & \ul{\textbf{0.71}} & \ul{{\color{gray}\textbf{2.14}}} & 125ms
    % <-- 3.5 times, 4 times 
    \\
    \textbf{DELO}
    & 1.30 & 2.97  & 2.19 & 11.99  & 1.71 & 4.88  & 1.58 & 3.34  & 7.42 & 2.42  & 1.00 & 2.17  & 1.01 & 2.58  & 1.44 & 1.97  & 3.48 & 9.02  & 1.54 & 2.26 
    % 4.26 (Actual)
    & 2.16 & 3.54 &  \ul{{\color{gray}\textbf{35}}}ms
    \\
    \textbf{DELO+PUEPE}
    & \ul{{\color{gray}\textbf{0.81}}} & \ul{{\color{gray} \textbf{1.43}}}  & \ul{{\color{gray}\textbf{0.57}}} & \ul{{\color{gray}\textbf{2.19}}}  & \ul{\textbf{0.52}} & \ul{\textbf{1.48}}  & \ul{{\color{gray}\textbf{1.10}}} & \ul{\textbf{1.38}}  & 1.70 & 2.45  & \ul{\textbf{0.64}} & \ul{\textbf{1.27}}  & \ul{\textbf{0.35}} & \ul{{\color{gray}\textbf{0.83}}}  & \ul{\textbf{0.41}} & \ul{\textbf{0.58}}  & \ul{\textbf{0.64}} & \ul{\textbf{1.36}}  & \ul{{\color{gray}\textbf{0.57}}} & \ul{{\color{gray}\textbf{1.23}}}  & \ul{{\color{gray}\textbf{0.90}}} & \ul{\textbf{1.53}} & \ul{{\color{gray}\textbf{41}}}ms 
    \\
    \bottomrule
    \end{tabular}
    \vspace{0.05cm}
    \caption{Results of different approaches for LiDAR odomety on KITTI \cite{Geiger2013IJRR} dataset are quantified by RRE, RTE metrics. The sequences \textbf{07-10} that are used for testing or inference, are \lq\textit{not seen}\rq~during training the network of the supervised approaches~\cite{Li_2019_CVPR,PWCLO-Net,WgICCV19,AliCVPR21} and ours.
    \\\hspace{\textwidth}{\ul{\textbf{Black}}}/\ul{{\color{gray}\textbf{Gray}}}: The best and second best entries are underlined and marked in bold \ul{\textbf{black}} and \ul{{\color{gray}\textbf{gray}}} color.
    \\\hspace{\textwidth}
    $\dagger:$ Denotes the error metrics are reported from from~\cite{PWCLO-Net}
    \\\hspace{\textwidth}
    PWCLO$^x$: The superscript $x$ means ($x\times1024$) number of input points are used for ~\cite{PWCLO-Net}.}
    \label{tab:kitti_Performance_Eval}
\end{table*}
\vspace{-0.2cm}
\subsection{Dataset, Baselines, and Evaluation Metrics}\label{DELO:Experiments_Evaluation:Baselines}
\vspace{-0.1cm}
\noindent\textbf{Dataset.~} We randomly select $70\%$ and $30\%$ of frame ids from the sequences 00-06 as source frames $f$. These frames are independent and mutually exclusive from each other. Therefore, we set the corresponding target frames with ids $f+o$. The ground truth transformations $\small \mathbf{T}_{gt}^{f}=\mathbf{T}_{f+o}^{-1}\mathbf{T}_{f}$ for every frame $f$ is set to the Velodyne coordinate system.     

\vspace{0.05cm}
\noindent\textbf{Baselines.~}\label{DELO:Experiments_Evaluation:Metrics} We evaluate our proposed approach against standout baseline methods -- PWCLO \cite{PWCLO-Net} network, 
LONet \cite{Li_2019_CVPR}, 
RPSRNet\cite{AliCVPR21},
DGR \cite{ChoyCVPR20},
DCP \cite{WgICCV19}, 
LOAM \cite{ZhangAR16} without mapping (\ie, no further scan-to-map alignment), 
ICP \cite{BlMcTPAMI92}, and 
ICP$_{\perp}$ \cite{segal2009generalized}. 
For evaluation, DCP, DGR, PWCLO network, and RPSRNet are re-trained. The POT-LDM and PU-EPE networks of DELO are jointly trained for 75 epochs with 1024 randomly selected points per scan, batch size of 16, and learning rate of 10$^{-4}$ on two NVIDIA GeForce 1080Ti GPUs.

\vspace{0.1cm}
\noindent\textbf{Evaluation Metrics.~}\label{DELO:Experiments_Evaluation:Metrics} The angular deviation $\mathbf{\varphi}$ between the ground truth and the predicted rotations ($\mathbf{R}_{gt}, \mathbf{R}$) and the relative distance error $\Delta t$ between the translation components ($\mathbf{t}_{gt}, \mathbf{t}$) 
\vspace{-0.1cm}
\begin{equation}\label{eqn:Angular_Trans_error}
\small
 \mathbf{\varphi} = \frac{180^{\circ}}{\pi} \cos^{-1}\left(
 0.5(\text{tr}\left(\mathbf{R}_{gt}^T\mathbf{R}\right) - 1)\right),\,\, \text{and}\,\, 
 \Delta t = \lVert\mathbf{t}_{gt} - \mathbf{t}\rVert
\end{equation}
quantify registration errors.  
On the other hand, the standard metrics to compare LO drifts are average of relative translation errors (RTE) and relative rotational errors (RRE) over all possible frames within the path lengths 100, 200, \ldots, 800\,m: RTE $t_\text{rel}$ in $\tfrac{\text{m}}{100\,\text{m}}$ as percentage $\%$, RRE $r_\text{rel}$  in $\tfrac{\text{degree}}{100\,\text{m}}$. 
\vspace{-0.3cm}
\subsection{LiDAR Odometry Evaluation on KITTI}\label{Exp:LO_Evaluation}
\vspace{-0.2cm}
\begin{figure}[ht]
\centering
\includegraphics[width=0.99\linewidth]{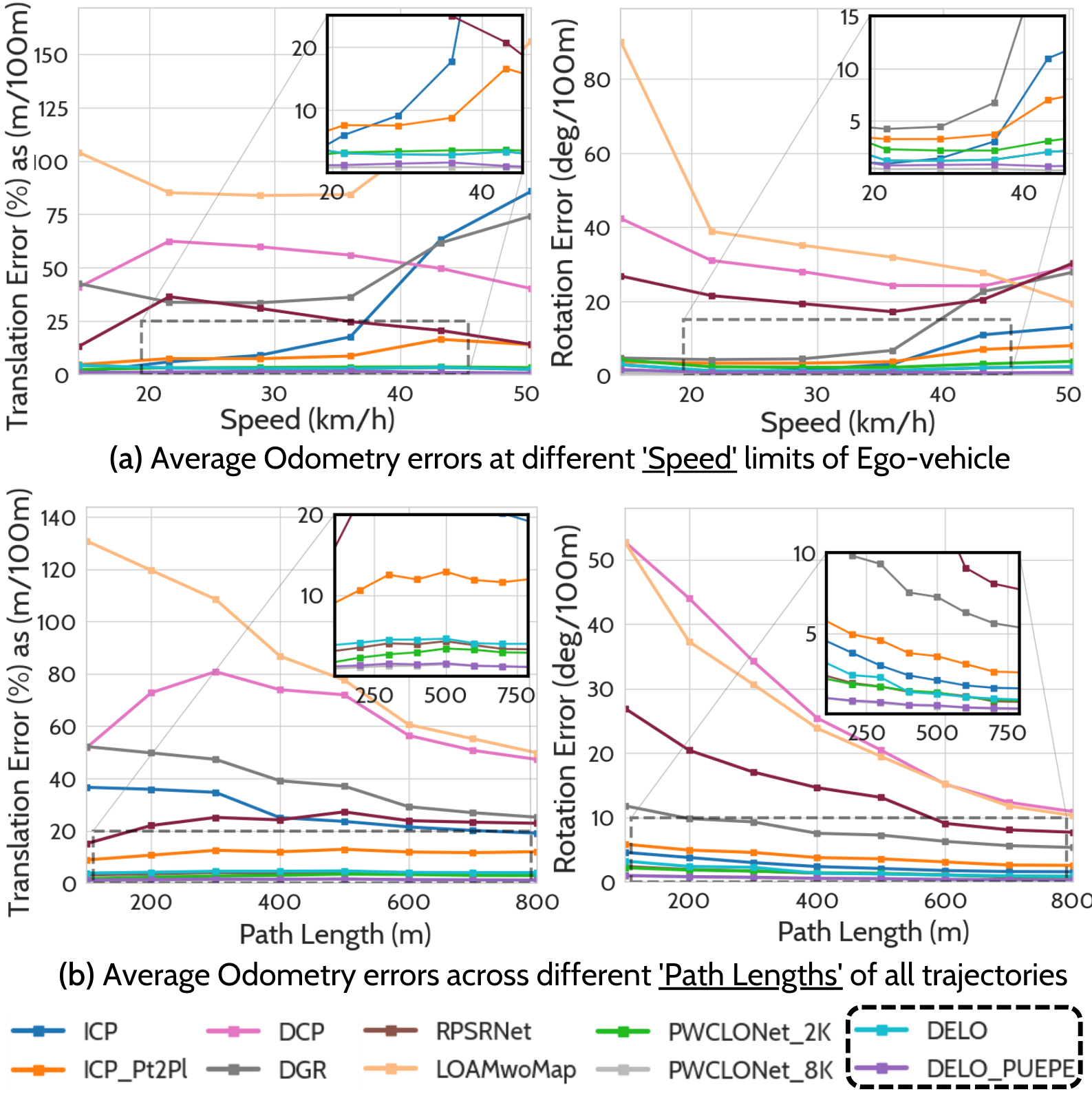}
\vspace{-0.1cm}
\caption{Relative transformation errors RRE and RTE averaged over all KITTI Odometry train and test sequences at different ranges of vehicle-speed (20Km/h, 30Km/h, ..., 50Km/h) and trajectory length (100m, 200m,...,800m). Our DELO+PUEPE performs the best.}
\label{fig:DELO_avg_error_seg_speed}
\end{figure}
\begin{figure*}
\includegraphics[width=\textwidth]{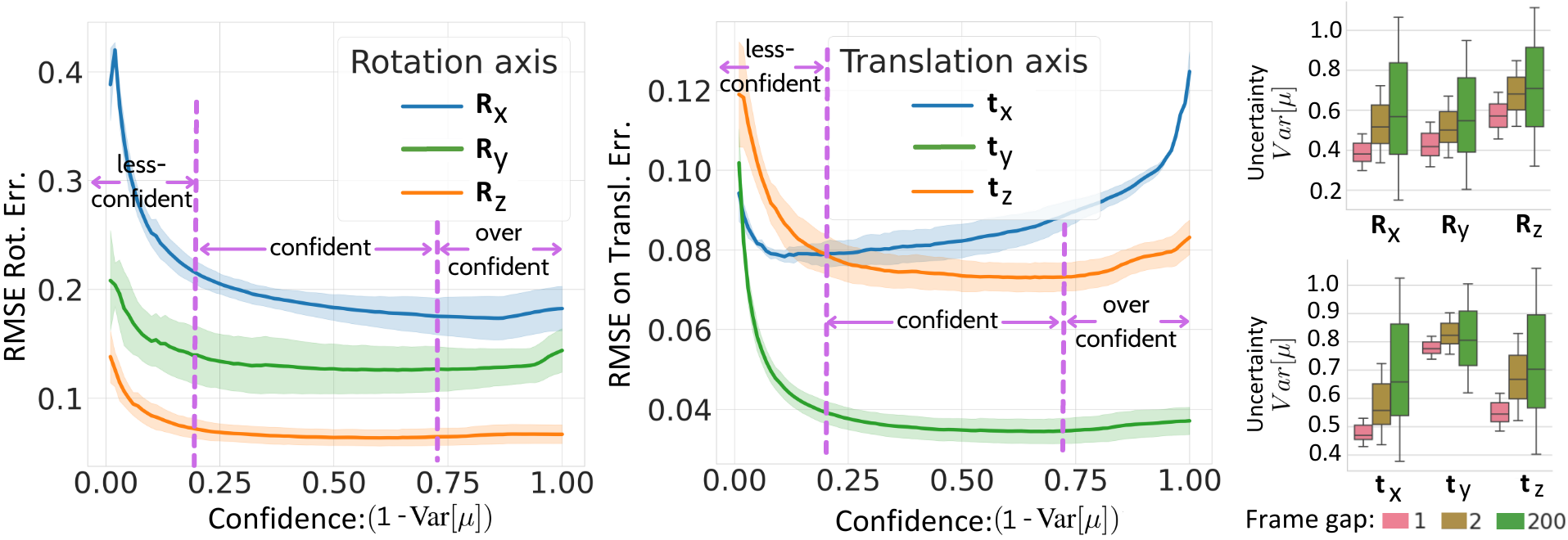}
\caption{\textit{On the left:} Root mean squared error (RMSE) along the transformation axes ($\mathbf{R}_x, \mathbf{R}_y,\mathbf{R}_z, \mathbf{t}_x,\mathbf{t}_y,\mathbf{t}_z$) over all frames selected under confidence cut-off percentiles of our DELO+PUEPE network. \textit{On the right:} Quartile plots over PU distribution along the transformation axes when different frame gaps are chosen for LO estimation.} \label{fig:Plot_Evidenntal_OdomError1}
\end{figure*}
%
%
% We thoroughly evaluate the performance of different baseline approaches. 
%
%
Table~\ref{tab:kitti_Performance_Eval} quantifies the most expressive error metrics~\cite{PWCLO-Net,Li_2019_CVPR,DeepLO_DMLO}, \ie,~RRE and RTE, for evaluating LO methods on every KITTI sequence. 
Lower values of the error tuple ($r_{\text{rel}},t_{\text{rel}}$) indicate stability of a method for a prolonged duration of continuous navigation. 
Therefore, the higher values of ($r_{\text{rel}},t_{\text{rel}}$) tuple reflect higher odometry errors due to drift accumulation along the trajectories. 
For a fair and consistent comparison, we split the training (including validation) and testing (or inference) set as Seq.~\textbf{00-06} and \textbf{07-10} as per \cite{PWCLO-Net,Li_2019_CVPR}. 
While training our network, we optimize the combined losses defined in Eq.~\eqref{eqn:total_Reg_Loss} and~\eqref{eqn:total_evidential_loss}.
To our understanding, LONet~\cite{Li_2019_CVPR} and PWCLONet~\cite{PWCLO-Net} are the two main benchmark methods to compete for performance superiority. %
Since the source codes LONet is not publicly available, we report its odometry errors (incl. ICP~\cite{BlMcTPAMI92}, ICP$_{\perp}$~\cite{segal2009generalized}, LOAM~\cite{zhang2014loam}, and PWCLONet~\cite{PWCLO-Net}) from~\cite{PWCLO-Net} in separate rows of Table~\ref{tab:kitti_Performance_Eval}.  
In another separate part, reported errors are from our experiments.

After evaluation, the first major observation is the drop in accuracy of the deep learning-based methods on the test sequences compared to the training sequences (last three rows of Table~\ref{tab:kitti_Performance_Eval}).
When there are only a handful of methods for deep learning-based LO, then the lack of their network's generalization ability on \lq\textit{unseen}\rq~data is a serious point to address. 
The proposed DELO, when trained with PU-EPE component, outperforms PWCLONet on three out of four test sequences. Although, on training sequences, PWCLONet outperforms our method only by marginal differences. 
Compared to the input size of 8K points in~\cite{PWCLO-Net}, DELO can accurately predict relative poses using matching correspondences between inputs of size 1K points per frame (see Fig.~\ref{fig:Ch6_MatchingMatrix_Comparison}). Furthermore, if PWCLONet is trained with randomly sampled 2K and 1K points as input, we observe a significant jump in its odometry errors across all sequences (see Table~\ref{tab:kitti_Performance_Eval}). In terms of runtime, after exceptionally fast RPSRNet and DCP, our method takes $\sim$30-41ms that can match scanning frequency of any modern LiDAR sensor. 
The next major observation is that state-of-the-art LiDAR data registration methods (\eg, DCP~\cite{WgICCV19}, DGR~\cite{ChoyCVPR20}, RPSRNet~\cite{AliCVPR21}) are not necessarily suitable for odometry task. 
For instance, on the test sequences, these methods score at least five times higher relative transformation errors than LONet~\cite{Li_2019_CVPR}, PWCLONet~\cite{PWCLO-Net}, and DELO+PUEPE (see Table~\ref{tab:kitti_Performance_Eval} and Figure~\ref{fig:DELO_avg_error_seg_speed}-(b)).
When the ego-vehicle changes its speed or drifts during navigation, relative motion between selected source and target frames appear out-of-order than the learned distribution of sensor motion. The Figure.~\ref{fig:DELO_avg_error_seg_speed}-(a) plots the relative transformation errors averaged over all possible frames from train and test sequences at different ranges of vehicle speed. The plots show our DELO, particularly DELO+PUEPE, and PWCLO-Net$^{8}$ methods are resistant to OOD. In contrast, other methods struggle to estimate correct relative motion when the ego-vehicle accelerates or decelerates.
%, \ie, when rolling and shutter occurs.  

%
%

%
%
Next, the performance of both point-to-point and point-to-plane ICP methods~\cite{BlMcTPAMI92,segal2009generalized} is derailed by continuous and catastrophic failures at different time steps. Similarly, methods like LOAM~\cite{zhang2014loam} or Lego-LOAM~\cite{ji2019lloam}, well-known for distorted LiDAR-SLAM, overly rely on extra scan-to-map alignment step. 
%
%
% Hence, without this step, their predictions are worse than ICP~\cite{segal2009generalized}.  
%
%

%
%
%\footnote{A detailed overview LO failures at different time steps of every KITTI sequence is available in the supplementary material.}.

%
%
\begin{figure*}[!ht]
\centering
\includegraphics[width=0.99\linewidth]{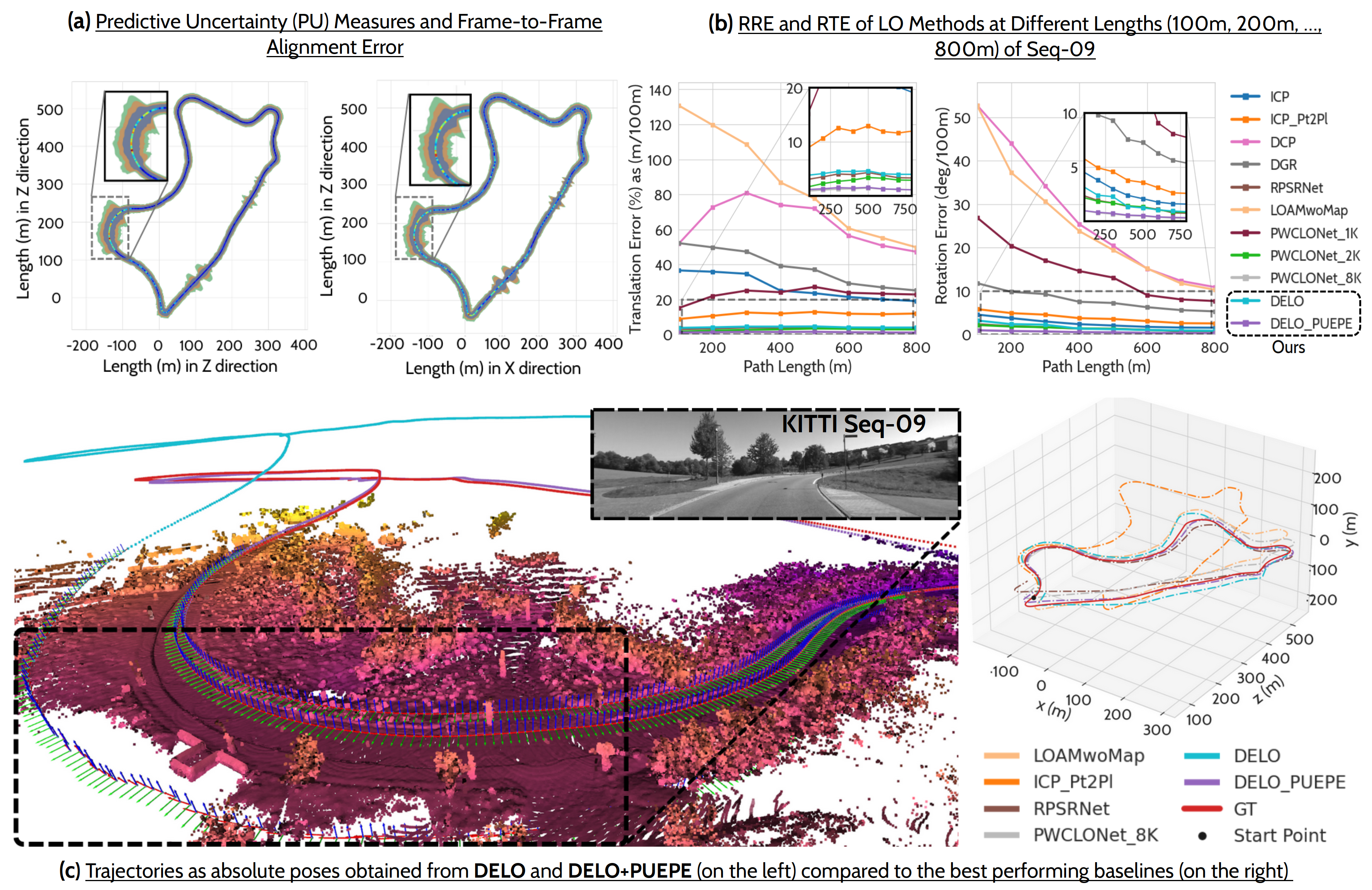}
\vspace{-0.1cm}
\caption{Complete experimental analysis on KITTI test Seq-09:
}
\label{fig:DELO_EvalAnalysis_Seq09}
\end{figure*}
\subsection{Equivariant PU-EPE as Evidence}\label{DELO:Experiments_Evaluation:PUEPE}
%\vspace{-0.15cm}
In the final part of our analysis, we demonstrate approximately equivariant nature of the epistemic uncertainty 
$Var[\mu] = \frac{\beta}{\nu (\alpha - 1)}$ w.r.t the odometry errors along all 6 directions (DoF) of the transformation parameters. The first two plots in Figure~\ref{fig:Plot_Evidenntal_OdomError1} show the RMSE on three Euler angles $\mathbf{R}_x, \mathbf{R}_y$, $\mathbf{R}_z$  for rotations, and three translation components $\mathbf{t}_x, \mathbf{t}_y$, $\mathbf{t}_z$ over all frames selected under the different cut-off confidence values (as opposite of uncertainty, \ie~$1 - \text{Var}[\mu]$). 
The transparent width of each line denotes how much the RMSE values vary by increasing odometry frame gap $o$ from 1 to 2. 
The last plot Figure~\ref{fig:Plot_Evidenntal_OdomError1} shows our model's uncertainty corresponding to different components of transformation axes, if different frame gaps (\ie, 1, 2, and 200) are chosen to predict odometry. 
The confidence percentiles and RMSE over all frames, selected by every cut-off percentile, intuitively classify three regions where -- DELO model is under-confident, confident, and over-confident. We set $\theta_{min}=0.2$ and $\theta_{max}=0.7$ in Eq.~\eqref{eqn:Confidence_Bound} using empirical analysis shown in Figure~\ref{fig:Plot_Evidenntal_OdomError1}.
Finally, Figure~\ref{fig:DELO_EvalAnalysis_Seq09} explains overall performance of our method using both the qualitative and quantitative results on the test sequence 09. Interestingly, the sequences 09 and 10 capture the same area via different routes, with narrow lanes covered by trees or bushes (see both the Figure~\ref{fig:DELO_EvalAnalysis_Seq09}-(c) and~\ref{fig:Ch6_DELO_Overview}-(b)). We plot the trajectory color-mapped by the POT-LDM prediction errors $\varphi$ and $\Delta t$ (see Eq.~\ref{eqn:Angular_Trans_error}), and the uncertainty values for $\mathbf{R}_x, \mathbf{R}_y$, $\mathbf{R}_z, \mathbf{t}_x, \mathbf{t}_y$, $\mathbf{t}_z$ as smooth 1D Gaussian filters along the same trajectory. It is noticeable in the Figure~\ref{fig:DELO_EvalAnalysis_Seq09}-(a), that there are clear evidences of continuous LO failures along the \lq\textit{circular turning point}\rq ~where transformation errors are high and DELO+PUEPE network signals either its over-confidence or under-confidence. After online pose-refinement, our method recovers the accurate absolute poses of ego-vehicle and performs the best among baseline approaches (See the RRE, RTE errors and 3D trajectory plots 
in Figure~\ref{fig:DELO_EvalAnalysis_Seq09}-(b) and (c)).   
\vspace{-0.25cm}
\section{Conclusions}
\vspace{-0.2cm}
This paper presents a real-time and LiDAR-only deep learning model for odometry estimation that jointly learns relative sensor motion and its predictive uncertainty. Our novel partial optimal transport network can learn sharp correspondence matching between two aggressively sub-sampled and non-uniformly distributed point clouds. The joint learning of odometry and its uncertainty leverages online pose-refinement by understanding under-confident or over-confident nature of predicted ego-motion. We show the equi-variant nature of PU across all transformation axes and all driving sequences can determine the thresholds for network to decide where sequential pose refinements are necessary. DELO is better than state-of-the-art methods, with great generalization ability, on KITTI dataset. 
%In future, we aim to extend our contributions towards multi-task learning approach for online SLAM, motion planning, and domain-adaption for LO.

\noindent\textbf{Acknowledgement.} This work was partially funded by the project DECODE (01IW21001) of the German Federal Ministry of Education
and Research (BMBF) and by the Luxembourg National Research Fund (FNR) under the project reference C21/IS/15965298/ELITE/Aouada.

{\small
\bibliographystyle{ieee_fullname}
\bibliography{egbib}
}

\end{document}